\documentclass[conference]{IEEEtran}
\IEEEoverridecommandlockouts
\usepackage{cite}
\usepackage{amsmath,amssymb,amsfonts}
\DeclareMathOperator*{\argmin}{argmin}
\usepackage{algorithmic}
\usepackage{graphicx}
\usepackage[dvipsnames]{xcolor}
\usepackage[final]{microtype}
\usepackage[italic]{mathastext}
\usepackage{libertine}
\usepackage[T1]{fontenc}
\usepackage{textcomp}
\usepackage[varqu,varl]{zi4}
\usepackage[all]{nowidow}
\usepackage{flushend}
\usepackage{fancyhdr}
\usepackage{enumitem}
\usepackage{mathtools, nccmath}
\usepackage{tablefootnote}
\usepackage{subcaption}
\usepackage{diagbox}
\usepackage{cite}
\usepackage{algorithm}
\usepackage[algo2e]{algorithm2e} 
\DeclarePairedDelimiter{\nint}\lfloor\rceil
\def\BibTeX{{\rm B\kern-.05em{\sc i\kern-.025em b}\kern-.08em
    T\kern-.1667em\lower.7ex\hbox{E}\kern-.125emX}}
\begin{document}

\title{Low-Bitwidth Floating Point Quantization for Efficient High-Quality Diffusion Models\\
}

\author{\IEEEauthorblockN{Cheng Chen}
\IEEEauthorblockA{\textit{Electrical and Computer Engineering} \\
\textit{University of Toronto}\\
Toronto, Canada \\
ccheng.chen@mail.utoronto.ca}
\and
\IEEEauthorblockN{Christina Giannoula}
\IEEEauthorblockA{\textit{Computer Science} \\
\textit{University of Toronto \& Vector Institute}\\
Toronto, Canada \\
cgiannoula@cs.toronto.edu}
\and
\IEEEauthorblockN{Andreas Moshovos}
\IEEEauthorblockA{\textit{Electrical and Computer Engineering} \\
\textit{University of Toronto \& Vector Institute}\\
Toronto, Canada  \\
moshovos@eecg.toronto.edu}
}

\maketitle

\begin{abstract}
Diffusion models are emerging models that generate images by iteratively denoising random Gaussian noise using deep neural networks. These models typically exhibit high computational and memory demands, necessitating effective post-training quantization for high-performance inference. Recent works propose low-bitwidth (e.g., 8-bit or 4-bit) quantization for diffusion models, however 4-bit integer quantization typically results in low-quality images. We observe that on several widely used hardware platforms, there is little or no difference in compute capability between floating-point and integer arithmetic operations of the same bitwidth (e.g., 8-bit or 4-bit).
Therefore, we propose an effective floating-point quantization method for diffusion models that provides better image quality compared to integer quantization methods. 
We employ a floating-point quantization method that was effective for other processing tasks, specifically computer vision and natural language tasks, and tailor it for diffusion models by integrating weight rounding learning during the mapping of the full-precision values to the quantized values in the quantization process. 
We comprehensively study integer and floating-point quantization methods in state-of-the-art diffusion models. Our floating-point quantization method not only generates higher-quality images than that of integer quantization methods, but also shows no noticeable degradation compared to full-precision models (32-bit floating-point), when both weights and activations are quantized to 8-bit floating-point values, while has minimal degradation with 4-bit weights and 8-bit activations. Additionally, we introduce a better methodology to evaluate quantization effects, highlighting shortcomings with existing output quality metrics and experimental methodologies. Finally, as an additional potential benefit, our floating-point quantization method increases model sparsity by an order of magnitude, enabling further optimization opportunities.
\end{abstract}


\section{Introduction}

 Diffusion models~\cite{song2021scorebased,ddim,ho2020denoising} have demonstrated remarkable success in image synthesis and image generation, surpassing the previous state-of-the-art GAN-based generative models\cite{dhariwal2021diffusion}. These models have proven to be highly effective in a variety of applications, including image superresolution~\cite{li2021srdiff}, inpainting\cite{lugmayr2022repaint}, restoration~\cite{li2023diffusion}, translation, editing~\cite{meng2022sdedit}, and even in autonomous vehicles~\cite{liu2024ddmlag,pronovost2023scenario}. 

Diffusion models generate new images by starting with random noise and using a noise estimation network, often the U-Net~\cite{ronneberger2015unet}, to iteratively denoise the input. The process begins with Gaussian-distributed random noise, which the network gradually denoises step-by-step until a final image is produced. Typically, tens to hundreds of denoising steps are used per image.

The denoising process of diffusion models is both compute- and memory-intensive, whether generating a single image or a batch of images. For example, the Stable Diffusion text-to-image model~\cite{rombach2022highresolution} uses a U-Net with 860 million parameters and typically requires 50 denoising steps per image. Generating one image with Stable Diffusion takes about 6 seconds on an Nvidia V100 GPU and 304 seconds on an Intel Xeon Gold 5115 CPU. This lengthy inference process makes it impractical to use Stable Diffusion in real-time applications deployed on resource-constrained edge devices.

Quantization~\cite{nagel2021white} reduces the memory footprint of neural networks by changing the representation of their data values from 32-bit floating-point (FP32) to narrower bitwidths. Quantization entails a trade-off between increased efficiency and lower output quality. Approaches to mitigate the impact of quantization on output quality are Quantization Aware Training (QAT) and Post Training Quantization (PTQ). QAT trains the network for a target data type~\cite{nagel2021white}, requiring a training dataset and significant time; for instance, training the Latent Diffusion Model-4 (LDM4) for one epoch takes about 7 hours on an NVIDIA A100~\cite{rombach2022highresolution}. QAT setup is also complex and may need refinement, if the target data type is too restrictive. PTQ, in contrast, starts with an already-trained network, does not need a training dataset, and is much faster, optionally using fine-tuning to reduce the impact.

Quantization poses unique challenges for diffusion models due to: 1) the noise introduced by quantization, and 2) multiple denoising steps during inference. The precision loss from quantization appears as noise in computed values. For models with a single pass, like image classification, this is less concerning. However, in diffusion models, the iterative denoising process during inference can cause quantization noise to accumulate and amplify over multiple time steps.

The use of integer PTQ without fine-tuning to reduce inference costs in diffusion models has received significant attention~\cite{shang2023ptqdm, li2023qdiffusion, he2023ptqd}. Recent methods~\cite{shang2023ptqdm, li2023qdiffusion, he2023ptqd} improve the balance between quantization aggressiveness and output quality, however further improvements are still needed. Using state-of-the-art integer PTQ methods, the generated image quality degrades, when quantizing to 8-bit integer (INT8), and the degradation is even greater with 4-bit integer (INT4) quantization.

In contrast to prior works, we argue that quantizing diffusion models to 8-bit floating point (FP8) instead of INT8, and to 4-bit floating point (FP4) instead of INT4, can be similarly highly advantageous in performance efficiency for two reasons. First, the memory footprint and bandwidth remain identical, as they depend on the bitwidth, not the bit interpretation. Second, on some platforms, there is little or no difference in compute throughput between integer and floating-point arithmetic operations. For example, this holds for desktop/server GPUs (2000 TFLOPS peak FP8 tensor core VS. 2000 TOPS peak INT8 tensor core on NVIDIA H100)~\cite{h100_indepth} and embedded Orin NVidia GPUs~\cite{orin}. Additionally, NVIDIA H100 GPU Tensor Core units have added support for FP8 format, providing twice the computational throughput of 16-bit operations~\cite{blog_nvidia}. Recently, NVIDIA announced Blackwell which has also added support for FP4~\cite{blackwell}.

However, it is not clear if floating-point quantization will provide better output quality over integer quantization. Careful consideration needs to be given, since although the bitwidth is the same, quantization to floating-point vs. to integer entails a non-trivial trade-off between precision and range: for the same bitwidth, the integer representation provides more precision for some values, while the floating-point provides a wider range. 

Therefore, our \emph{goal} in this work is to study floating-point quantization in diffusion models, targeting to improve output quality over the integer quantization with the same bitwidth. We quantize both weights and activations using a PTQ method. To our knowledge, our work is the \emph{first} to apply floating point quantization in diffusion models, and effectively quantize the weights of diffusion models to FP4.


We propose a floating point quantization method for diffusion models inspired by the method developed by Kuzmin et al.~\cite{kuzmin2024fp8}, which quantizes computer vision and natural language processing models. This method has not been applied to diffusion models, thus important improvements are needed, such that the method is effective for low-bitwidth quantized diffusion models. Specifically, we introduce a rounding learning technique (inspired by~\cite{nagel2020down}) in floating-point diffusion model quantization:  we leverage the gradient descent algorithm to learn how to map full-precision values (32-bit) to quantized values (8-bit or 4-bit) more effectively, rather than simply mapping full-precision values to their corresponding nearest values in the quantized range. This is the \emph{first} work that applies this key technique to floating-point quantization of deep neural networks.

We conduct a comprehensive evaluation of our floating-point quantization method versus prior integer quantization methods on two tasks: unconditional generation and text-to-image generation. In unconditional generation, we use the Latent Diffusion Model pre-trained on the LSUN-Bedrooms256$\times$256 dataset~\cite{yu2016lsun} and the DDIM~\cite{ddim} pre-trained on the 32$\times$32 CIFAR-10 dataset~\cite{cifar10}. These models are trained to generate a specific class of images. In text-to-image generation, we use Stable Diffusion pre-trained on the 512x512 LAION-5B datasets~\cite{schuhmann2022laion5b}, that generates different classes of images 
based on a textual description of what the image should depict. For our target quantization bitwidth, we quantize weights to both FP8 and FP4 and activations to FP8. 

We highlight the following experimental findings:

\begin{itemize} [noitemsep,topsep=0pt,nolistsep,leftmargin=10pt]
\item Our floating-point quantization method for diffusion models outperforms integer quantization at the same bitwidth, providing better quality. Since the memory costs and compute throughput of integer and floating-point representations of the same bitwidth are similar on many platforms, our method offers an attractive trade-off between efficiency and output quality.
\item In text-to-image generation, models with weights quantized to FP4 and activations to FP8 generate higher-quality images compared to models with both weights and activations quantized to INT8. 
\item We find that popular metrics used to measure the output quality of diffusion models do not always accurately reflect human perception of image quality, and should be used with caution. Visual inspection of integer-quantized model outputs reveals discrepancies between metric-reported quality and actual perceived quality.
\item Our quantization method increases the sparsity in weights of diffusion models by an order of magnitude. Sparsity can be exploited to further improve inference latency, since it provides additional optimization opportunities.
\end{itemize}



Overall, we make the following contributions:
    \begin{itemize}[noitemsep,topsep=0pt,nolistsep,leftmargin=10pt,nosep]
        \item We characterize the memory and compute requirements of stable diffusion for text-to-image generation, and we are the first to perform floating-point quantization in diffusion models.
        \item We propose a floating point quantization method that 
        uses (i) a greedy-search approach to assign per tensor floating-point formats, and (ii) a gradient-based rounding learning approach to improve output quality. Our work is the first to enable high output quality, 
        when quantizing weights and activations of diffusion models to FP4 and FP8. 
        \item We conduct the first experimental analysis comparing floating-point and integer quantization on diffusion models, and demonstrate that our method significantly improves output quality over prior approaches. 
        \item We introduce a better methodology for measuring output quality by choosing more appropriate reference images for fair comparisons. Additionally, our method increases sparsity in diffusion models, enabling further optimizations.
    \end{itemize}

\section{Diffusion Model Basics}
\begin{figure}[h]
            \centerline{\includegraphics[scale=0.052]{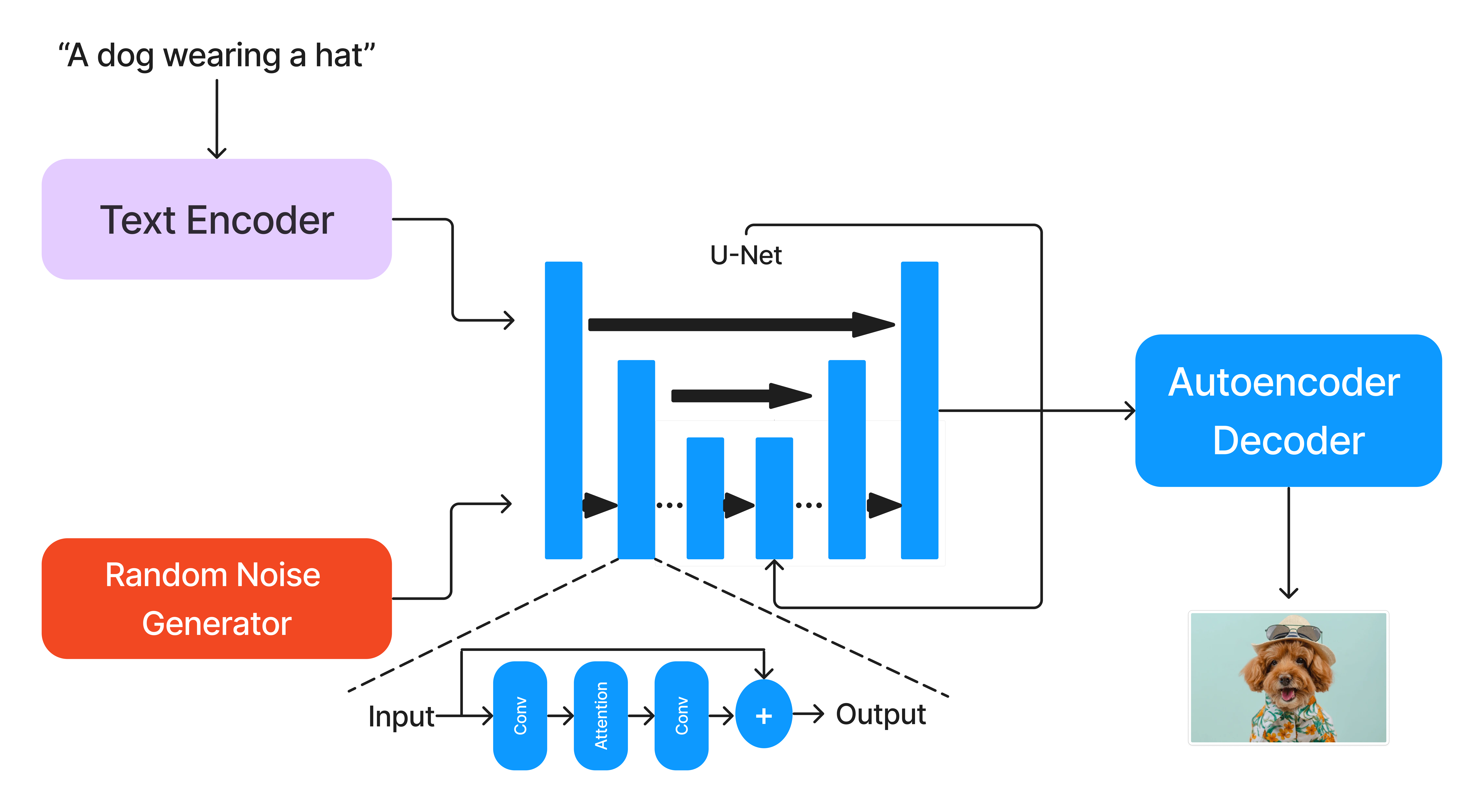}}
            \caption{Stable Diffusion Architecture.}
            \label{sd_visual}
        \end{figure}

        Diffusion models can perform a multitude of tasks. In this work, we focus on two widely used applications: unconditional generation and text-to-image generation. Figure~\ref{sd_visual} illustrates a high-level view of the Stable Diffusion text-to-image generation model~\cite{rombach2022highresolution}. Stable Diffusion accepts as input a Gaussian-distributed random noise and a prompt that describes the image to be generated. The prompt is then encoded by a ``text encoder''
        before getting passed to the U-Net denoising subnetwork. The U-Net then iteratively removes the noise from the noise input over multiple steps (typically 50). The final generated image is produced by the ``Autoencoder/Decoder'' subnetwork, which is invoked once at the end. The ``Autoencoder/Decoder'' converts the latent space output produced by the U-Net to the pixel space as the output image. 
        
         The U-Net itself comprises several ResNet~\cite{he2015deep} and Attention~\cite{vaswani2023attention} blocks. Figure~\ref{sd_visual} shows how the blocks are connected along with a simplified yet representative view of the internals of such a block. A typical block uses a combination of attention, convolutional, linear, group and/or layer normalization layers.  A distinctive characteristic of U-Net compared to other deep neural network architectures is the presence of block-to-block skip connections. A skip connection saves the activations from an earlier layer so that it can later concatenate them with the activations passed as input to another later layer. This technique has been proven to be helpful to the model's predictive or generative performance\cite{he2015deep}. Specifically, the output of the first block of the U-Net is also passed as input to the last block as shown by the arrows in Figure~\ref{sd_visual}. The output of the second U-Net block is passed as input to the last block, and so on. 
         This is \emph{not common} in other conventional neural networks such as transformers, where activations are typically consumed immediately after each block.

\begin{figure}[h]
            \centerline{\includegraphics[scale=0.14]{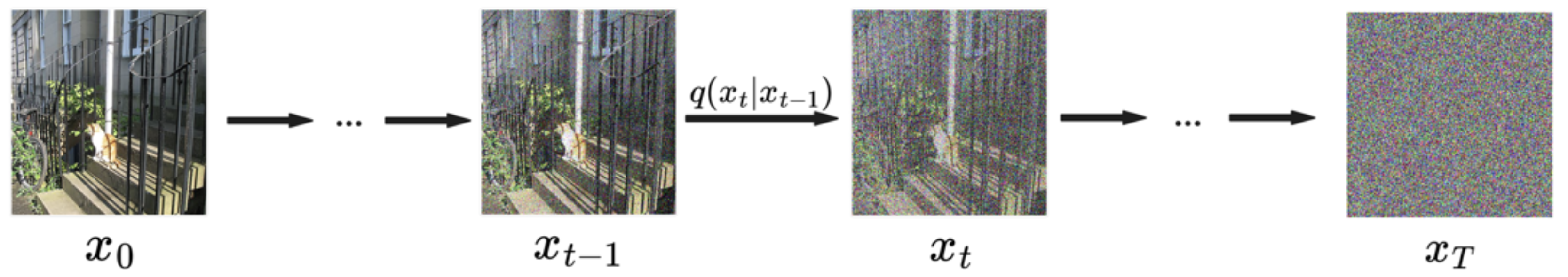}}
            \caption{Diffusion model forward process: it converts an image into Gaussian random noise.}
            \label{forward}
        \end{figure}

       Supervised training of diffusion models, as with any other neural network, requires a dataset with ground truth. To learn how to generate images from random noise, this dataset is generated as follows:  the \textit{forward process} starts with real data $x_0$ and incrementally introduces Gaussian noise to $x_0$. The Gaussian noise is added $T$ times, and if $T$ is sufficiently large, the \textit{forward process} ends up with random Gaussian noise. The \textit{forward process} is illustrated in Figure~\ref{forward}. Since the exact noise added at each step is known, the noise will be used as the ground truth during the training of the diffusion model. The forward process is defined as follows~\cite{ho2020denoising}:
            \begin{gather}
                q(x_{1:T} | x_0) = \Pi_{t=1}^{T}q(x_t | x_{t-1})\\    
                q(x_t | x_{t-1}) = \mathcal{N}(x_t;\sqrt{\alpha_t}x_{t-1},\beta_tI)
            \end{gather}
        where $\beta_t$ and $\alpha_t$ are hyperparameters regulating the intensity of Gaussian noise added at each step, $\beta_t$ = 1-$\alpha_t$. $x_t$ refers to the image at the current step, and $q(x_t | x_{t-1})$ is the sampled Gaussian noise to be added at step $t$.

\begin{figure}[h]
            \centerline{\includegraphics[scale=0.14]{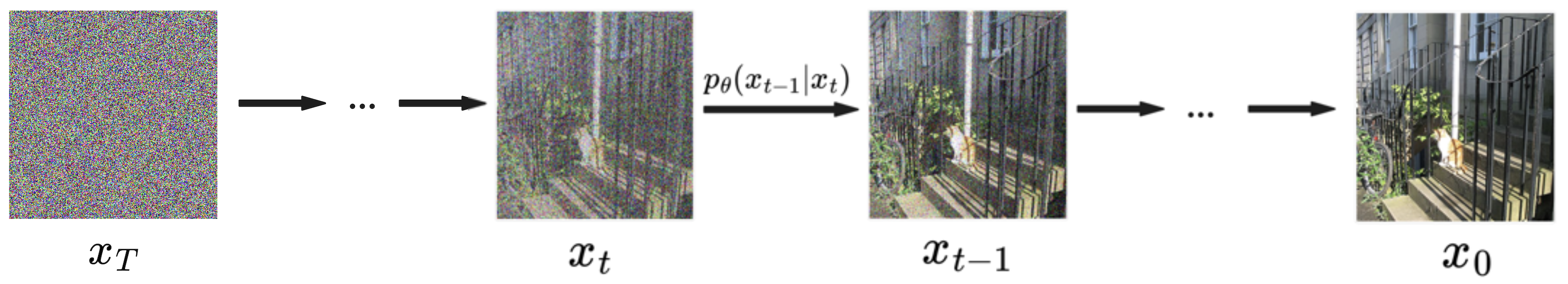}}
            \caption{Diffusion model backward process: it generates an image by iteratively denoising from a Gaussian random noise.}
            \label{backward}
            \vspace{-10pt}
        \end{figure}

        The \textit{backward process} aims to eliminate the noise from the noisy data to generate high-quality images. As shown in Figure~\ref{backward}, the backward process starts with random Gaussian noise $x_T$. The diffusion model eliminates the predicted noise from the noisy data at each step and repeats the process for $T$ steps. As the distribution $q(x_{t-1} | x_t)$ is intractable, diffusion models sample from a learned Gaussian distribution $p_\theta(x_{t-1}|x_t) = \mathcal{N}(x_{t-1};\mu_\theta(x_t,t),\Sigma_\theta(x_t,t))$, where the mean is reparameterized by a noise prediction network $\epsilon_\theta(x_t,t)$:
        \begin{gather}
            \mu_\theta(x_t,t) = \frac{1}{\sqrt{\alpha_t}}(x_t - \frac{\beta_t}{\sqrt{1-\bar{\alpha}_t}}\epsilon_\theta(x_t,t))
        \end{gather}
        where $\bar{\alpha}_t = \Pi_{i=1}^{t}\alpha_i$. 
        
        A detailed explanation of diffusion models and of their mathematical foundation is provided by Luo~\cite{luo2022understanding}.
        
\section{Characterization of Compute and Memory Requirements}
    In this section, we characterize Stable Diffusion during inference. We choose to characterize Stable Diffusion as text-to-image generation is a more challenging task compared to unconditional generation. Our analysis considers inference latency and the peak VRAM usage for GPU inference and motivates the need for quantization methods to reduce compute and memory costs.

    As mentioned, the U-Net is expected to handle most of the work in Stable Diffusion, since it runs multiple times during a single inference. We ran Stable Diffusion with a batch size of 1 and 50 denoising steps on an NVIDIA V100 (32GB VRAM, PyTorch~\cite{paszke2017automatic} version 1.12.1). We measure the runtime by running 12 times in total and averaging the last 10 runs. The U-Net accounts for 6.1 out of the total 6.6 seconds on average, while the text encoder and the autoencoder decoder together account for only 0.5 seconds.

\begin{figure}[h]
        \centerline{\includegraphics[scale=0.32]{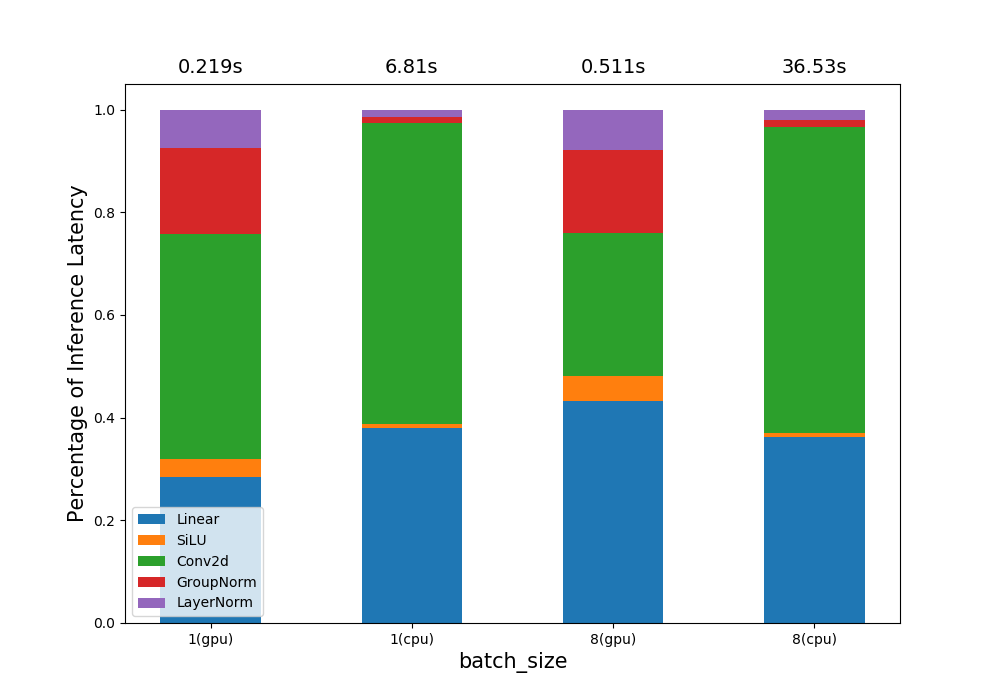}}
        \caption{{Breakdown} of inference latency for different types of layers, when running {Stable Diffusion with} batch size 1 and 8 on a CPU and a GPU.}
        \label{inf_time}
        \vspace{-10pt}
    \end{figure}

    Figure~\ref{inf_time} shows a breakdown of inference latency across the various layers in the U-Net. Latency is measured for one denoising step using the U-Net of Stable Diffusion and for batch sizes of 1 and 8 images in order to show the differences between smaller batch sizes and larger batch sizes. The figure shows the layer breakdowns on two types of machines: a CPU (Intel(R) Xeon(R) Gold 5115 @ 2.40GHz on CentOS-7 with 98GB RAM) and a GPU (NVIDIA Tesla V100 32GB VRAM on Red Hat Enterprise Linux 8.6). In either case, we use PyTorch~\cite{paszke2017automatic} version 1.12.1. The breakdown latencies are normalized to 1.0 to highlight the relative contribution of each layer type, with total inference latency in seconds shown above each bar.

    Most of the time is spent on the Conv2d and linear layers (including layers inside the attention units). The normalization and SiLU layers account for only about 25\% of the overall latency on the GPU, and a negligible portion on the CPU. The relative contribution across layer types changes little with the batch size on the CPU, whereas on the GPU, increasing the batch size to 8 increases the relative contributions of the linear layers. This is because increasing the batch size increases the memory traffic. Linear layers, which process larger arrays, are affected more by the increased batch size than conv layers, which process smaller arrays because conv layers use a kernel to sweep through the data. Since CPUs have smaller compute capability than GPUs, therefore they are compute-bound in most deep learning workloads, they show no significant change in relative latency between linear and conv layers with increased batch size. On the other hand, GPUs have higher compute capability than CPUs by parallelizing the workload, making GPUs more sensitive to increases in memory traffic compared to CPUs. Additionally, inference on the GPU is 31$\times$ and 72$\times$ faster than on the CPU for batch sizes of 1 and 8, respectively.
    
\begin{figure}[h]
        \centerline{\includegraphics[scale=0.37]{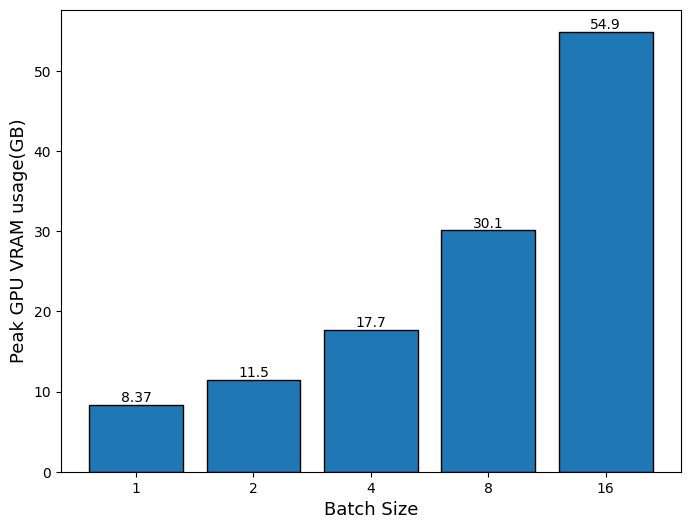}}
        \caption{Inference memory requirements}
        \label{inf_mem}
        \vspace{-10pt}
    \end{figure}

    Figure~\ref{inf_mem} reports the peak inference VRAM usage. The measurements were collected using Nvidia's Nsight Systems~\cite{nsightsys} on an Nvidia A100 80GB VRAM.  With a batch size of 16, the VRAM usage reaches as high as 54.9GB during inference out of the 80GB available on the specific GPU model. However, this is far beyond what is available on consumer level and embedded devices. Even if a single image is generated (batch size of 1), the memory consumption stands at 8.37GB. 
    
    Further investigation reveals that most of the memory consumed is largely due to the size of the data generated in the attention layers. In an attention layer, there are three linear layers that output the key, query, and values. The attention tensor is created by multiplying the key and query tensors. For example, the shape of the key and query tensor is (256,4096,40), when the batch size is 16, which results in a (256,4096,4096) tensor after multiplication. This tensor would require at least 17GB of memory, if the precision used is FP32. This VRAM requirement could be reduced by 4$\times$ and 8$\times$ by quantizing data values to FP8 and FP4, respectively.

        \section{Post-Training Quantization (PTQ)}
         Typically, models are trained to work with 32-bit floating-point. We will refer to these models as the \textit{baseline} or \textit{full-precision}. The goal of quantization is to reduce memory and/or compute costs by representing values with more efficient data types while, ideally, maintaining the output quality of the full-precision model. More efficient data types typically are those of lower precision, e.g., quantizing a model that originally used 32-bit  floating-point values to 8-bit integer values. 
         We will use the term ``\textit{task performance}'' to refer to any metric that quantifies the quality of the output.

         In post-training quantization (PTQ), the full-precision model is converted directly to a lower bitwidth model without retraining. Some PTQ methods use fine-tuning~\cite{10.1007/978-3-031-25082-8_8}, involving additional training steps. The dataset for fine-tuning may differ from the training dataset.
         In this work, we focus on PTQ methods that do not use fine-tuning, which would be a significant overhead in time and effort and would require access to an appropriate dataset.

         PTQ methods typically require only a lightweight calibration dataset to determine the quantization range, minimizing the error between the quantized and full-precision data values. In the next section, we will discuss methods for determining the quantization range. Generally, lower-bitwidth quantization results in greater quality degradation compared to the full-precision model.

         Quantization requires ``rounding'' some values of the original full-precision model to another value. The simplest method is rounding to the nearest representable value in the quantized model, but this is not always optimal. Recent studies propose alternative rounding strategies~\cite{nagel2020down} and block-wise reconstruction~\cite{li2021brecq} to mitigate quantization effects. However, these methods focus only on integer quantization, leaving low-bitwidth floating-point quantization relatively unexplored. Our goal is to investigate whether and to what extent narrow floating-point quantization can reduce the negative impact on task performance for diffusion models.

        Before we present our quantization method we first overview baseline integer and floating-point quantization approaches. 
        
    \subsection{Uniform Integer Quantization}\label{int_quant}       
        In uniform integer quantization, given a floating-point vector $X$ and the target integer bitwidth $b$, the quantization process is defined as: 
        \begin{gather}
            \label{int quant}
            X^{int} = s \cdot (clamp(\nint*{\frac{X}{s}} + z; 0, 2^b - 1) - z)
        \end{gather}
        where $\nint*{\cdot}$ is the round-to-nearest function and $clamp(\cdot;min,max)$ is the element-wise clipping operation to ensure that the quantized values are within the range that can be represented by the target bitwidth. The scaling factor $s$ is defined as $\frac{max(X)-min(X)}{2^b-1}$, and, finally, the zero point $z$ is $-\nint*{\frac{min(X)}{s}}$.
    \subsection{Floating Point Quantization}
        A standard floating-point number is represented as follows: 
        \begin{gather}
            f = (-1)^s2^{p-b}(1+\frac{d_1}{2}+\frac{d_2}{2^2}+\cdots+\frac{d_m}{2^m})
        \end{gather}
        where $s \in \{0, 1\}$ is the sign bit, $d_i \in \{0,1\}$ is the $m$-bit mantissa, $p$ is an integer  and $0 \leq p \leq 2^e$, where $e$ denotes the number of exponent bits, and $b$ is an  integer exponent bias, which is typically defined to be $2^{e-1}$.

        Floating-point numbers can be seen as a uniform $m$-bit grid between two consecutive integer powers: $2^a$, $2^{a+1}$. The distance between two adjacent grid points is $2^{a-m}$. Therefore, floating-point quantization is performed in similar fashion as described in (\ref{int quant}).

        The original values $X$ are first clipped to the range of the target floating-point format:
        \begin{gather}
            X' = clamp(X;-c,c)
        \end{gather}
        \begin{gather}
            \label{c_to_b}
            c = (2-2^{-m})2^{2^e-b-1}
        \end{gather}
        where $c$ is the absolute maximum value that the target floating-point format can represent.

        Changing the value of the exponent bias $b$  can control the range of the target floating-point number format. Then, quantization is performed on the clipped values: 
        \begin{gather}
            \label{fp_round}
            X^{fp}_i = clamp(s_i \nint*{\frac{X'_i}{s_i}};-c,c)
        \end{gather}
        where $X'_i$ denotes the $i$-th element of $X'$, and $X^{fp}_i$ denotes quantized $X'_i$. $s_i$ is the scale for the $i$-th element of $X'$ and is equal to:
        \begin{gather}
        s_i = 
            \begin{cases}
                2^{\lfloor\log_2 |X'_i| + b \rfloor - b - m} & \lfloor\log_2 |X'_i| + b\rfloor > 1\\
                2^{1-b-m} & otherwise.
            \end{cases}
        \end{gather}
        Kuzmin et al. present a more comprehensive overview~\cite{kuzmin2024fp8}.
        
        Unlike integer quantization, floating-point quantization allows for different encodings within the same bitwidth. In FP8, one bit is allocated for the sign, and the remaining 7 bits are divided between the exponent and mantissa bits. More bits can be allocated to the exponent vs. the mantissa if a wider range over precision is needed. In our study, we consider four candidate encodings for FP8: E2M5 (2-bit exponent and 5-bit mantissa), E3M4 (3-bit exponent and 4-bit mantissa), E4M3 (4-bit exponent and 3-bit mantissa), and E5M2 (5-bit exponent and 2-bit mantissa). For FP4, we consider two encodings: E1M2 (1-bit exponent and 2-bit mantissa) and E2M1 (2-bit exponent and 1-bit mantissa). Moreover, although the bias value is commonly defined as $2^{e-1}$, it can be modified. In this work, we use per tensor biases --- the bias value needs to be stored as metadata and using per tensor biases makes this metadata overhead negligible.

\section{Our Floating-Point Quantization Method}
    In this section, we present our method for floating-point quantization of diffusion models. As mentioned previously, the encodings and bias values can be different for each tensor in the same floating-point format. We start by discussing the method we use to choose the best per-tensor floating-point format (i.e., mantissa and exponent bit counts) and bias values.  Besides selecting the floating-point format and since the quality of generated images experienced significant degradation when quantized to FP4, we also apply rounding learning as Section~\ref{sec:rounding} explains. 
    \SetKw{KwBy}{by}
    \newcommand{\forconda}{$i=0$ \KwTo $len(encoding\_candidates)$}
    \newcommand{\forcondb}{$j=0$ \KwTo $len(bias\_candidates)$}
    \begin{algorithm}[H]
        \caption{Finding the bias value and encoding}\label{alg:search}
        \SetKwInOut{Input}{input}
        \SetKwInOut{Output}{output}
            
        \Input{$encoding\_candidates$, $bias\_candidates$}
        \Output{$encoding$,$bias$}
        
            $prev\_mse$  =  0\;\\
        \For{\forconda}{
            \For{\forcondb}{
                $curr\_encoding$ = $encoding\_candidates[i]$\; \newline
                $curr\_bias$ = $bias\_candidates[j]$\; \newline
                $curr\_mse$ = $get\_mse(curr\_encoding,curr\_bias)$\; \newline
                \eIf{$prev\_mse > curr\_mse$}{
                $encoding$ =  $curr\_encoding$ \; \\
                $bias$ = $curr\_bias$\;\\
                $prev\_mse$ = $curr\_mse$
                }{
                do nothing\;
                }
            }
    }
    \end{algorithm}

    \subsection{Encoding and Bias Value Selection}
    
   Before performing quantization, we need to select the encoding and the bias value for the target floating-point format. In floating-point representation, there are trade-offs between the number of bits allocated for the mantissa vs. the exponent. Increasing the mantissa bits enhances the precision of the values, while increasing the exponent bits expands the range of representable values.  To minimize the error introduced by quantization, it is crucial to carefully choose the encoding and bias value. 

   Kuzmin et al. \cite{kuzmin2024fp8} found that using different encodings and bias values for various quantized weight or activation tensors helps maintain task performance in image classification and natural language processing tasks. However, this method has not been tested on diffusion models. In this work, we adopt this method to investigate its effectiveness in quantizing diffusion models.
   
   Algorithm~\ref{alg:search} shows the search algorithm used to find the number of bits for the mantissa and the exponent, and also the bias per tensor. We consider both weight and activations. As we explain in detail below, since each tensor can have a different encoding including a tailored bias, we use a greedy approach to trim the search space. Specifically, the searching algorithm starts from the weight and activation tensors of the first layer in the model, selects the best possible choices for them, and fixes them before proceeding to the next layer. This process is repeated layer by layer (in a breadth first order) until all layers have been processed.

   Since the bias is a continuous real value, testing all possible values is impractical. Instead, we generate a set of evenly spaced values between the minimum and maximum of the data being quantized and calculate the bias for each based on (\ref{c_to_b}). We then use grid search to find the bias and encoding that minimize the mean squared error (MSE) between the quantized and full-precision data. For activation quantization, we use an \textit{initialization dataset}, a small sample from the full-precision model's output gathered uniformly across all timesteps (iterations of the U-Net), to obtain the full-precision data needed to calculate the MSE. Empirically, 111 bias values in the grid search provide the best trade-off between search time and task performance. Therefore, we use 111 bias value candidates for all our experiments.

    There are 4 encoding candidates for FP8 and 2 for FP4. Together with 111 bias values, there are 444 and 222 combinations for each weight tensor or activation tensor for FP8 and FP4 respectively, making it possible to search through each combination. For each combination of encoding and bias, we calculate the combination that achieves the lowest mean squared error between the quantized tensor and the full-precision tensor. 

    As Section~\ref{sec:eval} reports, this quantization method maintains task performance similar to the full-precision model, when using FP8. However, as Table~\ref{noRL_comp} shows, it fails to generate meaningful images with FP4 weights. In this experiment, weights are quantized to FP4 and activations to FP8. The table reports the \textit{Fréchet Inception Distance} (FID), a metric for generated image quality (discussed in Section~\ref{sec:eval}), where lower FID indicates better quality. The FID for FP4/FP8 models is 11.6$\times$ and 97.7$\times$ worse than the full-precision model for text-to-image and unconditional generation, respectively, indicating significant quality degradation. This result motivates the next step in our approach: applying a gradient-based rounding learning method to minimize quality loss, when quantizing weights to FP4.

    \begin{table}[t] 
        \caption{Ouput Image Quality Degradation with FP4 Weight/FP8 Activation Quantization}
        \begin{center}
        \begin{tabular}{c|c|c}
            \hline
              & Stable Diffusion & LDM(LSUNbedroom) \\
            \hline
            Bitwidth (W/A)$^1$ &FID$\downarrow$&FID$\downarrow$\\
            \hline
            Full Precision&22.71&2.95\\
            \hline
            FP4/FP8&262.8&288.2\\
            
        \label{noRL_comp}
        \end{tabular}\\
        \footnotesize{$^{1}$refers to the bitwidth quantization setting for weights and activations. For example, FP8/FP8 refers to when weights and activations are quantized to FP8}
        \vspace{-10pt}
        \end{center}
        \end{table}

    \subsection{Gradient-Based Rounding Learning for Low-Bitwidth Weights} \label{sec:rounding}
        In the context of low-bitwidth \textit{integer} quantization, Nagel et al.~\cite{nagel2020down} have shown that replacing the rounding-to-nearest operation with adaptive rounding can significantly improve task performance (see (\ref{int quant})). Since floating-point quantization also uses a rounding-to-nearest operation (see  (\ref{fp_round})), we adopt the learned rounding method previously proposed in the context of integer quantization to the floating-point domain and study whether it proves beneficial also in this context. The rest of this section describes how this approach \textit{learns} what rounding to use for each value in a tensor.
        
        Quantization can be seen as a lossy compression method. To mitigate the impact of quantization on task performance, one approach is to reduce the quantization error. Let's denote the quantized weights as $W^q$ and the original weights as $W$ and the output activations from the previous layer $A$. 
        
        The quantization error can be defined as follows:
        \begin{gather}
            \Delta Y = L(Y^q,Y) = L(W^qA,WA)
        \end{gather}
            where $Y^q$ represents the output of a layer in the quantized model, and $Y$ represents the output of the same layer in the full-precision model. In the above formula, we are considering a linear layer without bias as an example for ease of understanding. This method can be applied to other layers such as Convolutional and with having a bias. Following the approach of prior work~\cite{choukroun2019lowbit,kuzmin2024fp8}, we choose our objective as minimizing the MSE between the quantized layer output versus original layer output:
        \begin{gather}
            L(Y^q,Y) = L(W^qA,WA) = mean((W^qA-WA)^2) 
        \end{gather}

        To enable learning the rounding policy to be used for quantizing the weights, we modify (\ref{fp_round}) to:
        \begin{gather}
            W^q_i(\alpha_i) = clamp(s_i \cdot(\lfloor{\frac{W^q_i}{s_i}}\rfloor + \sigma(\alpha_i));-c,c)
        \end{gather}
        replacing the rounding-to-nearest operation with the floor operation $\lfloor{\cdot}\rfloor$ and adding the sigmoid function $\sigma(\alpha) \in [0,1]$, where $\alpha$ is the parameter we are optimizing using gradient descent. The sigmoid function is chosen, because its output is in the range [0,1], which simulates rounding down or rounding up. 
        
\begin{figure}[h]
            \centerline{\includegraphics[scale=0.5]{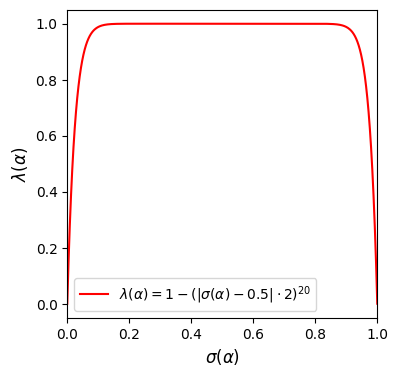}}
            \caption{Regularization term to push the output of the sigmoid function to the boundary of the interval [0,1] }
            \label{regularizer}
        \end{figure}

        The optimization objective then becomes: 
        \begin{gather}
            \argmin_\alpha mean((W^q(\alpha)A-WA)^2) + \lambda(\alpha)\\
            \lambda(\alpha) = 1 - (|\sigma(\alpha)-0.5|*2)^{20}
        \end{gather}
        where $\lambda(\alpha)$ is the regularization term visualized in Figure~\ref{regularizer}. This regularization term is minimized when $\sigma(\alpha)$ approaches 0 or 1, and therefore encourages $\sigma(\alpha)$ to converge towards 0 or 1 during rounding learning. During inference time, if $\sigma(\alpha)$ is less than 0.5, we set $\sigma(\alpha)$ to 0 indicating rounding down, and to 1 if $\sigma(\alpha)$ is greater than or equal to 0.5 indicating rounding up.

        To learn the rounding we use a \textit{calibration dataset}, which we generate as follows: we obtain full-precision data used to calculate the loss during rounding learning, we save several samples from the output of each full-precision model timestep and then randomly select a few samples from the collected data as the calibration dataset for each iteration of rounding learning.
        
        Overall, our FP4 quantization starts with the initialization dataset to select the optimal encoding and bias value, followed by rounding learning using the calibration dataset to further recover the lost task performance from quantization.

        Since FP8 quantization already maintains full-precision model task performance  without rounding learning (Section~\ref{sec:eval}), we apply the rounding learning method only to FP4 quantization.

\section{Experiments and Results} \label{sec:eval}
    \subsection{Methodology}\label{eval:methodology}
        To evaluate our quantization methods, we conduct image synthesis experiments using state-of-the-art diffusion models: DDIM~\cite{ddim}, Latent Diffusion Model (LDM), Stable Diffusion~\cite{rombach2022highresolution} and Stable Diffusion XL (SDXL)~\cite{podell2023sdxlimprovinglatentdiffusion}. We evaluate the models on two tasks: unconditional generation and text-to-image generation. For unconditional generation, we use LDM trained on the LSUN-Bedrooms dataset (256×256)\cite{yu2016lsun} and DDIM trained on CIFAR-10~\cite{cifar10}, and generate 50,000 samples, following Q-diffusion\cite{li2023qdiffusion}. For text-to-image generation, we use Stable Diffusion V1.5, sampling 2,000 prompts from the MS-COCO dataset~\cite{lin2015microsoft} and generating 10,000 samples. For SDXL, we generate 2,000 samples for evaluation. We set 200 denoising steps for unconditional generation and 50 steps for text-to-image generation, consistent with the original LDM settings.
        
        
        To construct the initialization dataset for searching mantissa bits and bias, we uniformly sample 128 samples from all denoising timesteps for unconditional generation and 16 samples for text-to-image generation. To construct the calibration dataset for rounding learning, we sample 5 samples from each step for unconditional generation and 20 samples from each step for text-to-image generation.
  
        During each iteration of gradient-based rounding learning, we randomly select 16 samples for unconditional generation and 8 samples for text-to-image generation from the calibration dataset. The number of samples in the calibration dataset is chosen empirically by comparing the task performance of the quantized model using different calibration data sizes. To compare our method with integer quantization for diffusion models, we use Q-diffusion~\cite{li2023qdiffusion}, the state-of-the-art integer quantization method, instead of PTQD~\cite{he2023ptqd}, which focuses on quantization noise correction rather than the quantization method itself. Q-diffusion recommends separately quantizing the skip connection and the previous layer's output before concatenation due to their different value distributions. We apply this technique to floating-point quantization as well.
        
        In all our experiments, we first quantize the weights and then the activations, with quantization performed on a per-tensor basis. For both tasks, as most of the inference time is spent on convolution and linear layers, and following standard practice, we quantize the weights and activations of convolutional and linear layers and leave the normalization layers and Silu activation function running in full precision. This is exactly the same approach used by prior integer quantization methods for diffusion models.

         \begin{figure*}[htbp]
            \centering
            \begin{subfigure}[t]{0.20\textwidth}
                \centering
                \includegraphics[width=\textwidth]{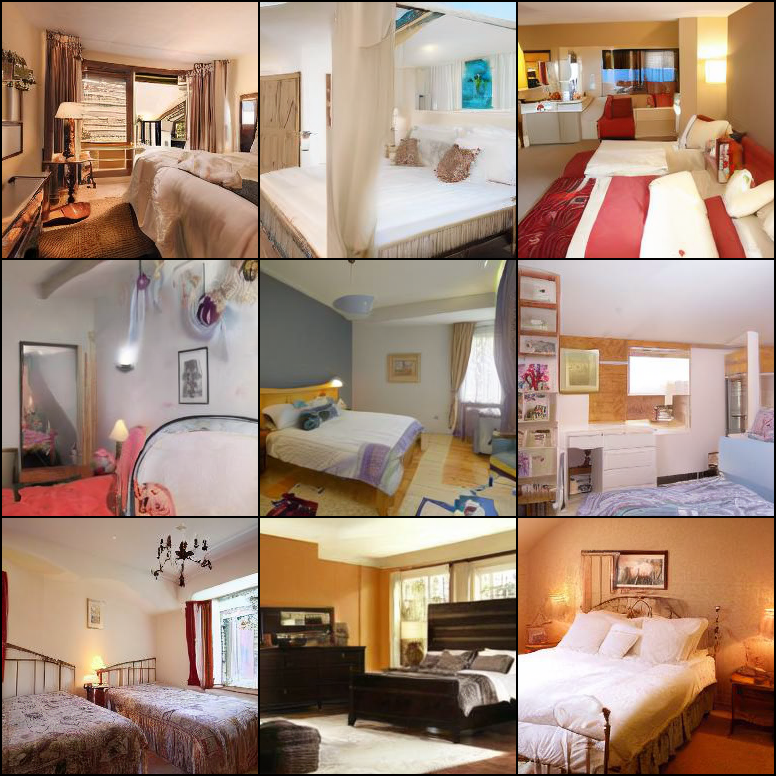}
                \caption{full-precision}
            \end{subfigure}
            \hfill
            \begin{subfigure}[t]{0.20\textwidth}
                \centering
                \includegraphics[width=\textwidth]{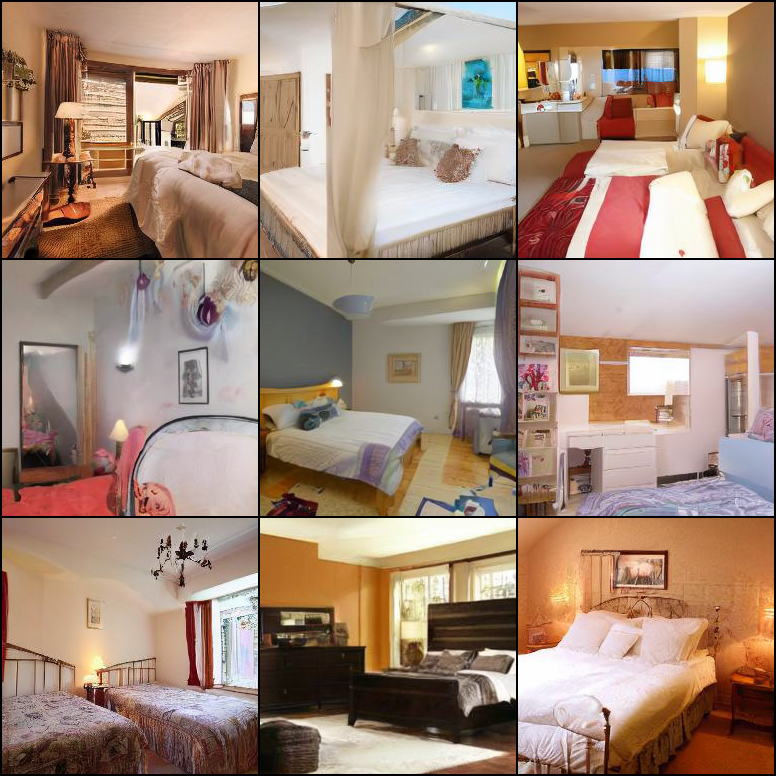}
                \caption{FP8/FP8}
            \end{subfigure}
            \hfill
            \begin{subfigure}[t]{0.20\textwidth}
                \centering
                \includegraphics[width=\textwidth]{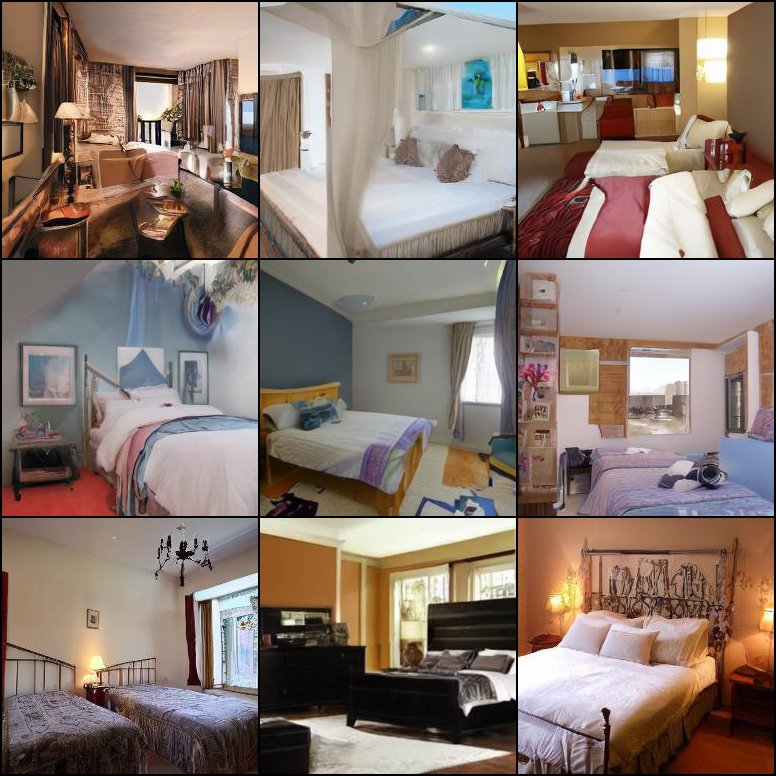}
                \caption{FP4/FP8}
            \end{subfigure}
            \hfill
            \begin{subfigure}[t]{0.20\textwidth}
                \centering
                \includegraphics[width=\textwidth]{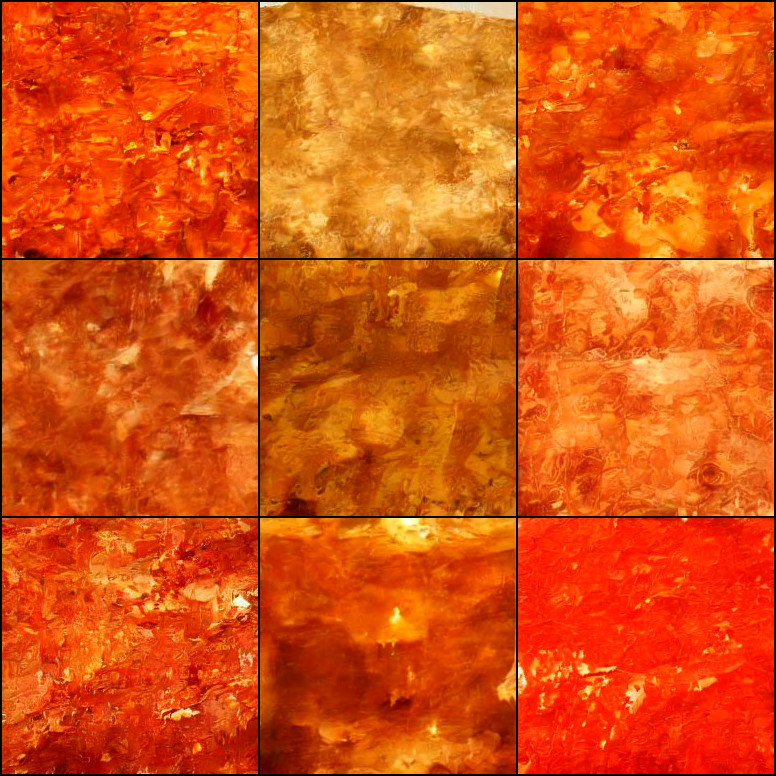}
                \caption{FP4/FP8 without rounding learning}
            \end{subfigure}
            \vspace{-10pt}
            \caption{LDM(LSUNbedroom) Qualitative Evaluation: example images generated by different quantized models}
            \label{bed_qualitative_eval}
            \vspace{-10pt}
        \end{figure*}
    \subsection{Image Generation Quality Metrics}\label{eval:metrics:quality}
        We report four metrics: Fréchet inception distance (FID)~\cite{heusel2018gans}, Spatial Fréchet Inception Distance (sFID)~\cite{nash2021generating}, Precision, and Recall~\cite{kynkäänniemi2019improved,sajjadi2018assessing}. For FID and sFID, we first extract features from the generated images and the reference images using Inception V3~\cite{szegedy2015rethinking}, and then compare the distance between the \textit{distributions} of the two sets of features. The difference between FID and sFID is that sFID uses spatial features rather than the standard pooled features, providing a sense of spatial distributional similarity between the images. Since this is a distance metric, lower values indicate better quality. 
        
        \textit{Precision} is the probability that a random image from the generated images falls within the support of the reference image distribution. \textit{Recall} is the probability that a random image from the reference images falls within the support of the generated image distribution. 
        
        For text-to-image generation, we also report the \textit{CLIP score}\cite{hessel2022clipscore}, which measures the similarity between the prompts and generated images. It is important for text-to-image generation that the generated images are both high-quality and relevant to the input prompts.

        We make the key observation that commonly used metrics do not always reflect changes in output image quality. To demonstrate this, we collect numerical metrics and visually inspect several output samples. There are cases where the metrics fail to fully describe the effects on quality because they use distributions as proxies and compare the distance between the distributions of reference and generated images. Often, quantization effects are not captured, when reference images differ significantly from the generated ones. We propose a better methodology for choosing reference images to more accurately evaluate the effects of quantization on output quality.

    \subsection{Facilitating Fair Comparisons Across Runs}
    Since the diffusion models use randomly generated noise as input, they naturally produce different images each time they are invoked. However, to facilitate proper comparisons across different experiments we fix the seed across runs that are to be compared. This way the input noisy data to the model remains the same for each batch at different runs. This ensures the generation of ``identical'' images for each batch, facilitating a meaningful comparison.  
    
    \subsection{Unconditional Image Generation}\label{eval:unconditional} 
        
        In this section, we evaluate our quantization methods on the LDM pre-trained on the 256$\times$256 LSUN-Bedroom dataset and DDIM pre-trained on the 32$\times$32 CIFAR-10 dataset, and compare against state-of-the-art integer quantization~\cite{li2023qdiffusion}. We use the same reference dataset used in Q-diffusion to measure all the metrics. 
        
        Quantitatively evaluating generative models is a challenging task, and it is possible that different metrics show different trends. Therefore, we include four metrics for a more comprehensive evaluation and to compare to Q-diffusion. The Q-diffusion study uses only FID for the evaluation of unconditional generation. 
        
        Table~\ref{cifar_eval} and Table~\ref{bed_eval} report the generated image quality metrics for different quantization methods and bitwidth targets. Recall that FP32/FP32 (weights/activations) is the baseline full-precision model. All other rows that use floating point were generated via our methods.  
        
       Table~\ref{cifar_eval} shows the generated image quality metrics for DDIM pre-trained on CIFAR-10. Both INT8/INT8 and FP8/FP8 models maintain the full-precision model performance, while there is minimal degradation when the weights are quantized to 4 bits. Although INT4/INT8 slightly outperforms FP4/FP8 in FID and Recall, FP4/FP8 is significantly better in sFID.
       
       Table~\ref{bed_eval} shows the generated image quality metrics for LDM pre-trained on LSUNBedroom dataset. Comparing FP8/FP8 with INT8/INT8 and FP4/FP8 with INT4/INT8 shows that the models quantized with our methods outperform the corresponding integer quantized models of the same bitwidths in three out of the four metrics. INT8/INT8 scores higher only in Recall, and INT4/INT8 scores higher in Precision. Regardless, both score differences are minor. Moreover, as we discuss below, inspection of several, randomly selected output images shows \emph{no} noticeable degradation of output quality.

        The FP8/FP8 model outperforms the full-precision model in FID and Precision (we will discuss this result further during the visual inspection of output images). In contrast, the INT8/INT8 model struggles to maintain the full-precision model's task performance, as measured by FID. This indicates that the FP8/FP8 quantized model does not degrade image quality, while reducing memory and compute costs by 4x. Furthermore, the rounding learning method is essential for FP4/FP8, as without it, all FP4/FP8 metrics indicate major failure, with Precision equal to 0.00 and Recall equal to 0.0146, suggesting that the generated images and reference images are significantly different. Later in this section, we show that these images are, in fact, close to random noise.

        \begin{table}[htbp]
            \caption{CIFAR10 Quantitative Evaluation}
            \begin{center}
            \begin{tabular}{c|c|c|c|c}
                \hline
                Bitwidth (W/A) & \textbf{\textit{FID$\downarrow$}}& \textbf{\textit{sFID$\downarrow$}}& \textbf{\textit{Precision$\uparrow$}} &\textbf{\textit{Recall$\uparrow$}} \\
                \hline
                Full Precision (FP32/FP32) &4.20&4.44&0.6657&0.5847\\
                \hline
                INT8/INT8&4.02&4.73&0.6406&0.5970\\
                FP8/FP8 (Ours)&3.70&4.31&0.6619&0.5954\\
                \hline
                INT4/INT8&4.67&5.94&0.6496&0.5820\\
                FP4/FP8 (Ours)&5.03&4.89&0.6513&0.5816
            \label{cifar_eval}
            \end{tabular}\\
            \vspace{-10pt}
            \end{center}
        \end{table}

        \begin{table}[htbp]
            \caption{LDM(LSUNBedroom) Quantitative Evaluation}
            \begin{center}
            \begin{tabular}{c|c|c|c|c}
                \hline
                Bitwidth (W/A) & \textbf{\textit{FID$\downarrow$}}& \textbf{\textit{sFID$\downarrow$}}& \textbf{\textit{Precision$\uparrow$}} &\textbf{\textit{Recall$\uparrow$}} \\
                \hline
                Full Precision (FP32/FP32) &2.95&7.05&0.6494&0.4754\\
                \hline
                INT8/INT8&3.29&7.51&0.6394&0.4806\\
                FP8/FP8 (Ours)&2.93&7.44&0.6559&0.4706\\
                \hline
                INT4/INT8&4.36&7.99&0.6598&0.4404\\
                FP4/FP8 no RL (Ours)$^1$ &288.21&151.96&0.00&0.0146\\
                FP4/FP8 (Ours)&3.84&7.36&0.6247&0.4742
            \label{bed_eval}
            \end{tabular}\\
            \footnotesize{$^{1}$refers to no rounding learning}
            \vspace{-10pt}
            \end{center}
        \end{table}

        Recall, that FID reports that FP8/FP8 is slightly better than the full-precision model. This is not impossible since it has been observed that in neural networks, sometimes a loss in precision can act as a regularizer defending against overfitting. Figure~\ref{bed_qualitative_eval} shows a few examples of the generated images which are representative of the random samples of output images we visually inspected. Our visual inspection of the images from the full-precision model (part~(a)) and from the FP8/FP8 model (part~(b)), reveals \emph{no} obvious change in image quality.  
        
        Considering images generated by the FP4/FP8 model (part~(c)), we observe that the color is not as bright as the images from the full-precision model or FP8/FP8 model, but the degradation in terms of the image composition is negligible. On the other hand, without applying the gradient-based rounding learning method on FP4 quantization, the quantized model fails to generate meaningful images (part~(d)).

        \begin{table*}[htbp]
        \caption{Stable Diffusion Quantitative Evaluation}
        \begin{center}
        \begin{tabular}{c|c|c|c|c||c|c|c|c}
            \hline
            \textbf{Reference Dataset}&\multicolumn{4}{|c||}{\textbf{MS-COCO}} & \multicolumn{4}{c}{\textbf{Full Precision Model Generated Images}}\\
            \hline
            \diagbox{Bitwidth (W/A)}{Metric} & \textbf{\textit{FID$\downarrow$}}& \textbf{\textit{sFID$\downarrow$}}& \textbf{\textit{Precision$\uparrow$}} & \textbf{\textit{Recall$\uparrow$}} & \textbf{\textit{FID$\downarrow$}}& \textbf{\textit{sFID$\downarrow$}}& \textbf{\textit{Precision$\uparrow$}} & \textbf{\textit{Recall$\uparrow$}} \\
            \hline
            Full Precision&22.71&64.81&0.6892&0.3966&0.00&0.00&1.00&1.00\\
            \hline
            INT8/INT8&21.29&63.55&0.6817&0.4230&5.53&33.38&0.8293&0.8659\\
            \hline
            FP8/FP8 (Ours)&22.23&64.17&0.6896&0.3924&2.76&18.50&0.9744&0.9746\\
            \hline
            INT4/INT8&20.95&63.54&0.6737&0.4313&5.85&33.97&0.8141&0.8474\\
            \hline
            FP4/FP8 no rounding learning (Ours) &262.63&266.53&0.007&0.00&253.25&220.53&0.0215&0.00\\
            FP4/FP8 (Ours) &21.75&64.93&0.6526&0.4232&5.53&31.73&0.8426&0.8925\\
            
        \label{SD_eval}
        \end{tabular}\\
        
        \end{center}
        \end{table*}

            \begin{figure*}[htbp]
            \centering
            \begin{minipage}{\textwidth}
            Prompt: A woman stands in the dining area at the table \\[0.5ex]
                \begin{minipage}{0.16\textwidth}
                    \centering
                    \includegraphics[width=\textwidth]{figs/mscoco_1.png}
                    \\[-0.5ex]
                \end{minipage}\hfill
                \begin{minipage}{0.16\textwidth}
                    \centering
                    \includegraphics[width=\textwidth]{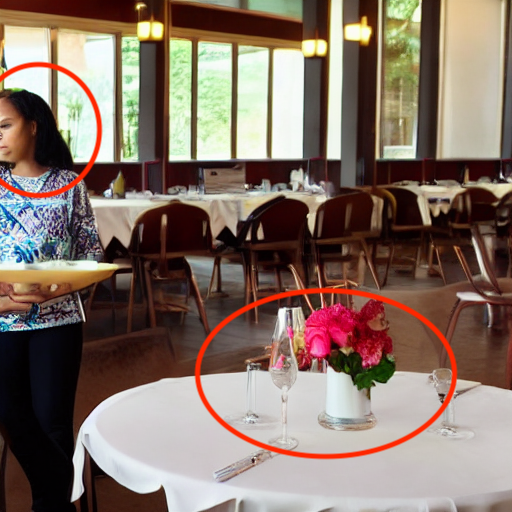}
                    \\[-0.5ex]
                \end{minipage}\hfill
                \begin{minipage}{0.16\textwidth}
                    \centering
                    \includegraphics[width=\textwidth]{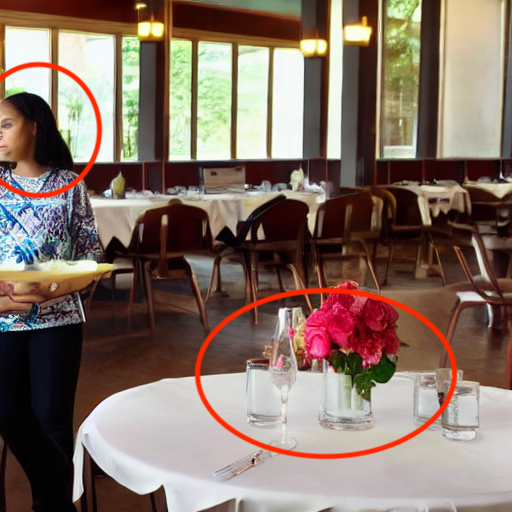}
                    \\[-0.5ex]
                \end{minipage}\hfill
                \begin{minipage}{0.16\textwidth}
                    \centering
                    \includegraphics[width=\textwidth]{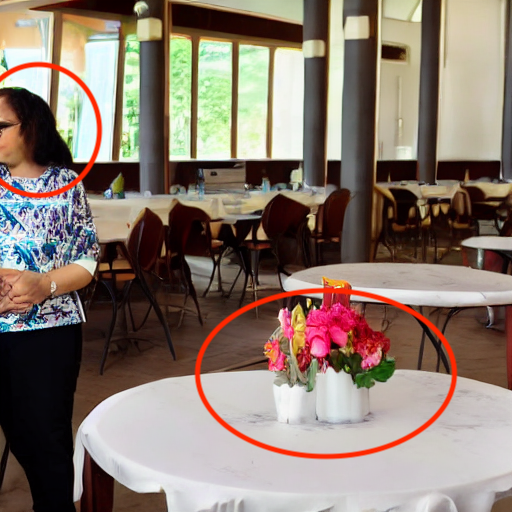}
                \end{minipage}\hfill
                \begin{minipage}{0.16\textwidth}
                    \centering
                    \includegraphics[width=\textwidth]{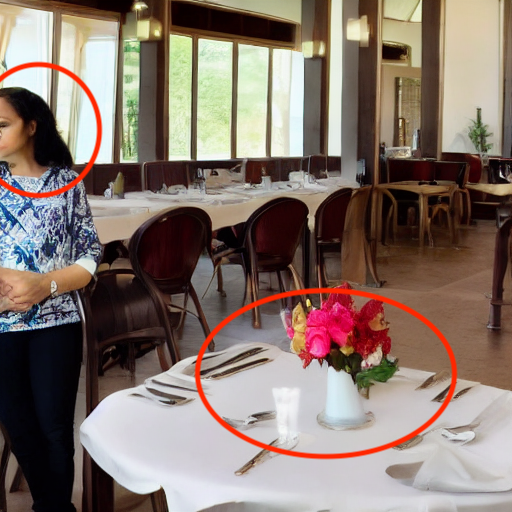}
                    \\[-0.5ex]
                \end{minipage}\hfill
                \begin{minipage}{0.16\textwidth}
                    \centering
                    \includegraphics[width=\textwidth]{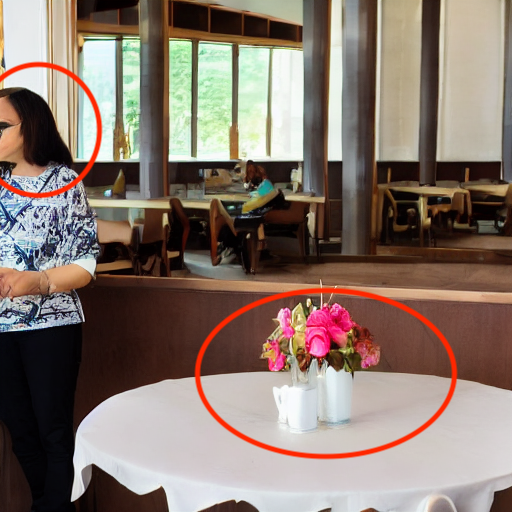}
                \end{minipage}
            \end{minipage}
            \\[\smallskipamount]
            \begin{minipage}{\textwidth}
            Prompt: Bedroom scene with a bookcase, blue comforter and window\\[0.5ex]
                \begin{minipage}{0.16\textwidth}
                    \centering
                    \includegraphics[width=\textwidth]{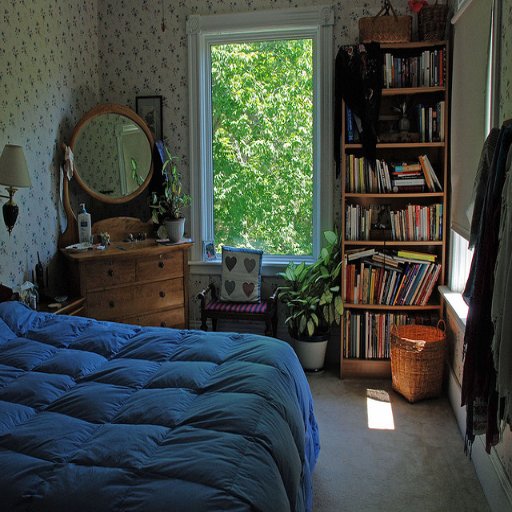}
                    \\[-0.5ex]
                    MS-COCO
                \end{minipage}\hfill
                \begin{minipage}{0.16\textwidth}
                    \centering
                    \includegraphics[width=\textwidth]{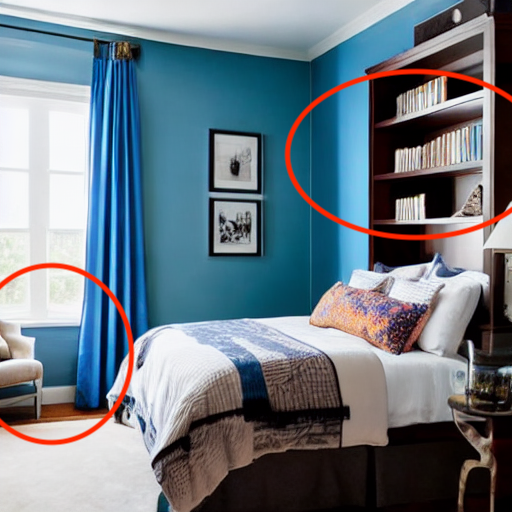}
                    \\[-0.5ex]
                    full-precision
                \end{minipage}\hfill
                \begin{minipage}{0.16\textwidth}
                    \centering
                    \includegraphics[width=\textwidth]{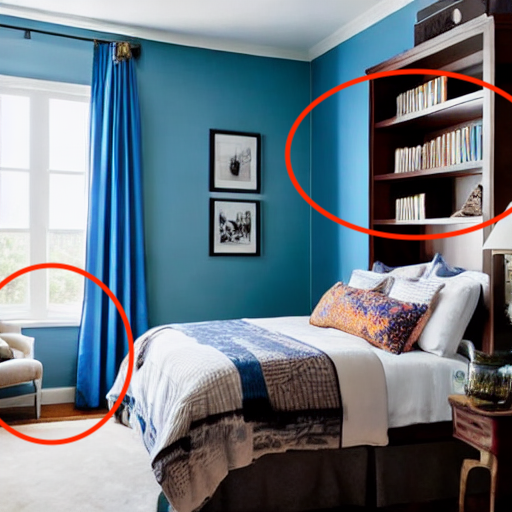}
                    \\[-0.5ex]
                    fp8/fp8
                \end{minipage}\hfill
                \begin{minipage}{0.16\textwidth}
                    \centering
                    \includegraphics[width=\textwidth]{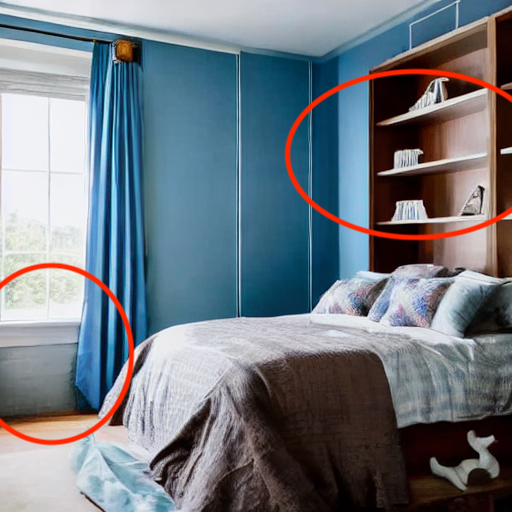}
                    \\[-0.5ex]
                    int8/int8
                \end{minipage}\hfill
                \begin{minipage}{0.16\textwidth}
                    \centering
                    \includegraphics[width=\textwidth]{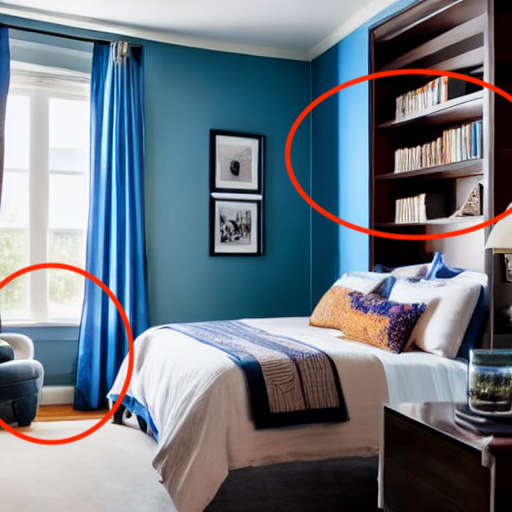}
                    \\[-0.5ex]
                    fp4/fp8
                \end{minipage}\hfill
                \begin{minipage}{0.16\textwidth}
                    \centering
                    \includegraphics[width=\textwidth]{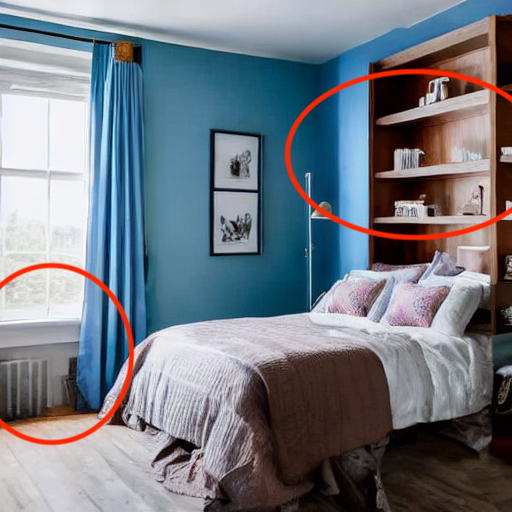}
                    \\[-0.5ex]
                    int4/int8
                \end{minipage}
            \end{minipage}
        
            \caption{Stable Diffusion Qualitative Evaluation}
            \label{SD_qualitative_eval}
            \vspace{-10pt}
        \end{figure*}

    \begin{table}[htbp]
            \caption{SDXL Quantitative Evaluation}
            \begin{center}
            \begin{tabular}{c|c|c|c|c}
                \hline
                \textbf{Reference Dataset}&\multicolumn{4}{|c}{\textbf{Full-precision Generated Images}}\\
                \hline
                Bitwidth (W/A) & \textbf{\textit{FID$\downarrow$}}& \textbf{\textit{sFID$\downarrow$}}& \textbf{\textit{Precision$\uparrow$}} &\textbf{\textit{Recall$\uparrow$}} \\
                \hline
                Full Precision&0.00&0.00&1.00&1.00\\
                INT8/INT8&94.22&247.42&0.135&0.681\\
                FP8/FP8 (Ours)&39.52&229.21&0.5125&0.894\\
            \label{sdxl_eval}
            \end{tabular}\\
            \vspace{-10pt}
            \end{center}
        \end{table}

        \begin{figure*}[htbp]
            \centering
            \begin{subfigure}[t]{0.29\textwidth}
                \centering
                \includegraphics[width=\textwidth]{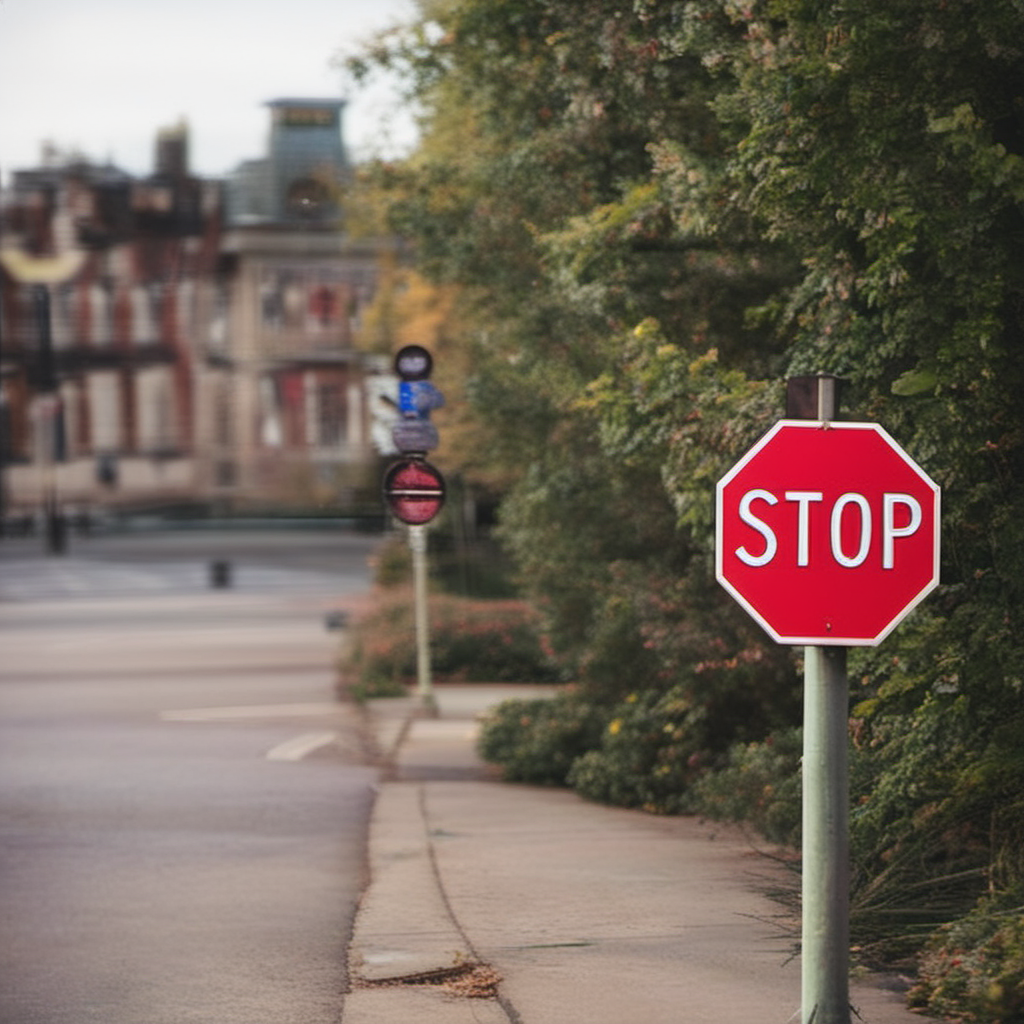}
                \caption{full-precision}
            \end{subfigure}
            \hfill
            \begin{subfigure}[t]{0.28\textwidth}
                \centering
                \includegraphics[width=\textwidth]{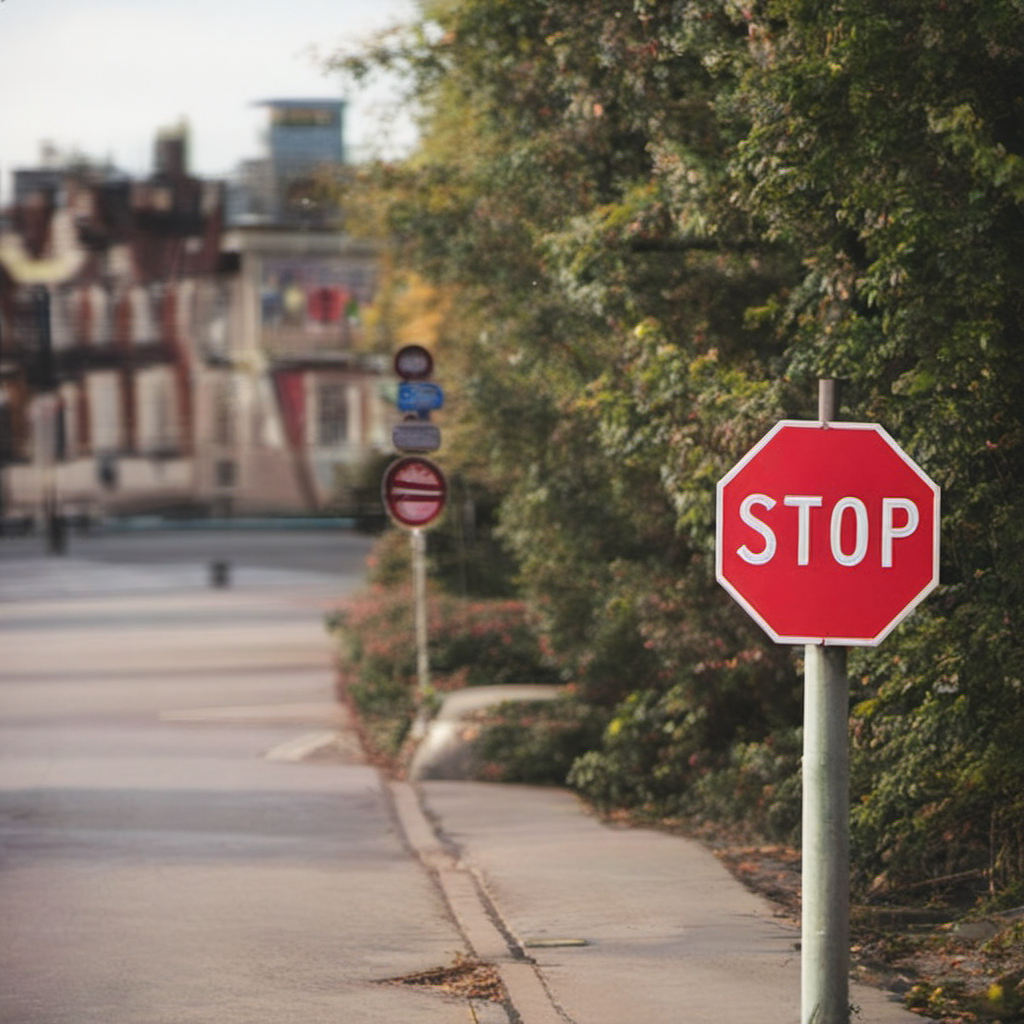}
                \caption{fp8/fp8}
            \end{subfigure}
            \hfill
            \begin{subfigure}[t]{0.28\textwidth}
                \centering
                \includegraphics[width=\textwidth]{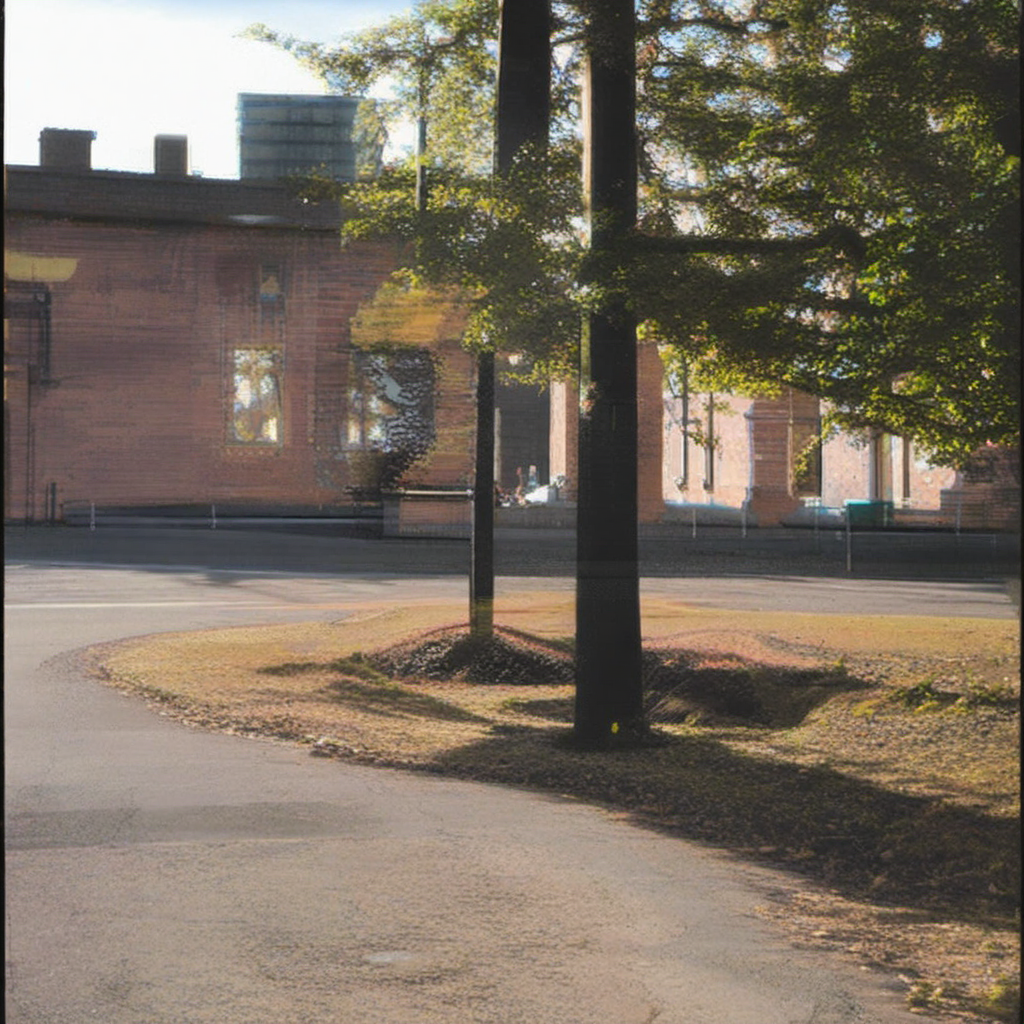}
                \caption{int8/int8}
            \end{subfigure}
            \hfill
            \caption{SDXL Qualitative Evaluation}
            \label{sdxl_qualitative_eval}
            \vspace{-10pt}
        \end{figure*}
        
    \subsection{Text-to-Image Generation}\label{sec:text2image}

        This section evaluates our quantization method on Stable Diffusion and SDXL pre-trained on the 512$\times$512 LAION-5B dataset. We calculate the output quality metrics using the MS-COCO 2017 validation dataset as reference images following the methodology of Q-diffusion. As shown in Table \ref{SD_eval}, both integer and floating-point quantized models maintain performance of the full precision model as reported by the metrics, with integer quantization achieving lower FID and sFID scores compared to floating-point quantization at the same quantization setting.      
        Further, as we quantize the models to lower bitwidths for both integer and floating-point quantization, the metrics show improvements compared to higher bitwidth settings. 

        Contrary to what the metrics suggest, qualitative inspection reveals that image quality degrades with lower bitwidth in integer quantization. Floating-point quantized models produce better-quality images, closer to those generated by the full-precision model. Figure~\ref{SD_qualitative_eval} shows several examples illustrating the poor quality from integer quantization. In the first row, faces generated by integer quantized models are blurry, and table utensils have disappeared. In contrast, floating-point quantized models maintain facial features and table details. In the second row, integer models fail to produce physically meaningful books on the bookcase and the chair beside the bed has vanished.
        Moreover, the INT4/INT8-generated images show less diversity compared to INT8/INT8-generated ones. For example, in the first row, the chairs and tables have disappeared in the background. 

        The above observations suggest that the current methodology used to evaluate the effect of quantization on output quality ought to be revisited. Below we present what we believe is a better methodology for doing so.
        
        \noindent\textbf{A Better Methodology To Measure Output Quality: } Without the loss of generality we focus our discussion on the FID metric. 
        
        The FID metric's goal is to measure how similar two \textit{sets} of images are. In our application, FID is meant to compare the quality between a reference ``ground-truth'' \textit{set} of images vs. a sample of the \textit{set} of images the model produces. That is, in FID we do not compare a single reference image vs. another produced image that is supposed to be as identical as possible. 
        FID converts each of the image sets into a distribution and then compares these distributions. The assumption is that the distributions are good enough proxies for the image content. 
               
        We first note that the images generated by the full-precision model differ significantly from the reference images from MS-COCO. This should have been expected since Stable Diffusion was trained on the LAION-5B dataset~\cite{schuhmann2022laion5b} and not on MS-COCO. Using a data distribution as a representative summary of the image set is less meaningful in this context given that the image sets differ by design.       
        
        The MS-COCO contains real-world images and was used as reference images in the original work that proposed Stable Diffusion to evaluate whether the generated images are close to real-world images. Since we are not making any modifications to the full-precision model architecture and since the goal of quantization is to reduce memory and compute costs while maintaining full-precision model task performance, we propose to use images generated by the full-precision model as the reference set, when evaluating the effects caused by quantization.
        
        In this revised approach, Table \ref{SD_eval} shows that, when using the same bitwidth, floating-point quantization outperforms integer quantization in all metrics. The FP8/FP8 model's FID is 2.77 lower (better) vs. the INT8/INT8, whereas the FP4/FP8 model's FID is 0.32 lower vs. the INT4/INT8. Notably, FP4/FP8 achieves the same FID, and better sFID, Precision and Recall compared to INT8/INT8. Similarly to unconditional generation, the rounding learning method for FP4 weight quantization reduces the degradation significantly as all 4 metrics for FP4/FP8 without rounding learning are significantly larger than FP4/FP4 with rounding learning.

        SDXL is the largest open-source image generative model, with a U-Net that is approximately three times larger than that of Stable Diffusion. Table~\ref{sdxl_eval} presents the image generation quality metrics for FP8 and INT8 quantized models. The FP8/FP8 model significantly outperforms the INT8/INT8 model across all four metrics. Figure~\ref{sdxl_qualitative_eval} provides an example of images generated by each model. The image generated by the FP8/FP8 model closely resembles the one produced by the full-precision model. In contrast, the image generated by the INT8/INT8 model is vastly different and lacks many details (e.g., the stop sign is absent).
\subsection{CLIP Score} 
        Thus far, the metric reported measures the output image quality with respect to a reference image set. Since Stable Diffusion is a text-to-image model, it is crucial to also measure that the generated images closely align with the input prompts. The CLIP score is the best practice metric used for this purpose. 

\begin{figure}[h]
            \centerline{\includegraphics[scale=0.47]{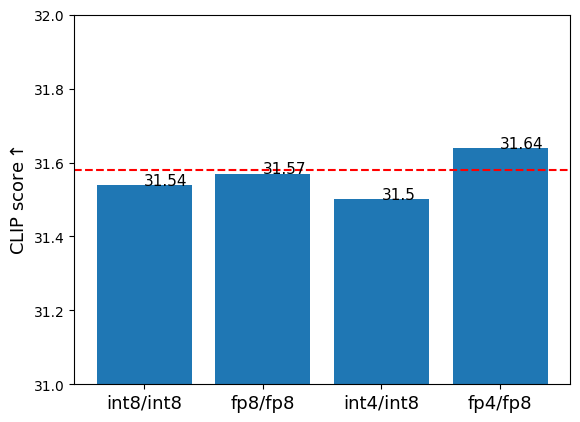}}
            \caption{Stable Diffusion CLIP Score. The dotted red line indicates the CLIP score of images generated by the full-precision model}
            \label{SD_clip}
        \end{figure}

        Figure~\ref{SD_clip} reports the CLIP score for various bitwidths. The differences across all methods are relatively small, even when compared to the full-precision model, suggesting that all models perform reasonably well according to the CLIP score. However, our FP8/FP8 and FP4/FP8 quantized models consistently exhibit better CLIP scores compared to integer quantized models, with the FP4/FP8 model achieving a slightly better CLIP score than the full-precision model.

    \subsection{Sparsity}  \label{sec:sparsity} 
        \begin{figure}[h]
            \centerline{\includegraphics[scale=0.47]{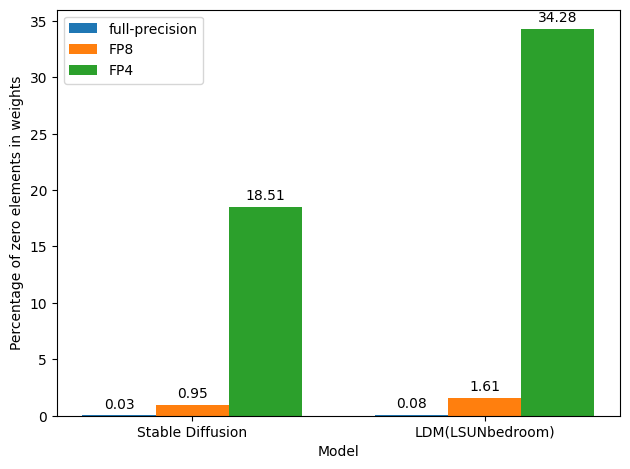}}
            \caption{Percentage of weights that are zero in Stable Diffusion and LDM(LSUNbedroom)}
            \vspace{-10pt}
            \label{sparsity}
        \end{figure}
    
    Since ultimately the goal of methods such as quantization is to reduce overall memory and compute costs, we also study the effect of quantization on another behavior that neural networks often exhibit, i.e., sparsity.  Sparsity, that is the fraction of values in a tensor that are zero, can be used to reduce memory footprint and transfers via appropriate encoding, and to reduce computations by skipping multiplications.
    
    Since quantization reduces data precision, it can naturally introduce sparsity by forcing values that are close to zero to become zero. Many previous works~\cite{10.1145/3597031.3597057,mishra2021accelerating,10.1145/3620665.3640356,10.1145/3410463.3414654,9164900} exploit sparsity in deep neural networks to reduce compute and memory costs. For example, NVidia GPUs via cuSCNN~\cite{10.1145/3597031.3597057} add hardware support for \textit{structural sparsity},  allowing for up to a certain pattern and number of zeros in subgroups of a tensor~\cite{mishra2021accelerating}. 
    
    We measure the sparsity in floating-point quantized models. Figure~\ref{sparsity} illustrates that for the weights of Stable Diffusion, our methods introduce a 31.6x and 617x increase in sparsity, when quantizing to FP8 and FP4, respectively, compared to the full-precision model. For LDM (LSUNBedroom), our methods introduce a 20.1$\times$ and 428.5$\times$ increase in sparsity, when quantized to FP8 and FP4, respectively, compared to the full-precision model. Leveraging the methods that exploit sparsity mentioned above, the inference latency and energy efficiency of diffusion models can be further improved together with our quantization methods.

\section{Related Work}
    A few recent works~\cite{li2023qdiffusion,shang2023ptqdm,he2023ptqd,qdm,so2023temporaldynamicquantizationdiffusion} have studied quantization methods in diffusion models, and their effects on the output quality of diffusion models.  PTQ4DM~\cite{shang2023ptqdm} was the first work to apply PTQ to quantize diffusion models trained on low-resolution datasets with INT8. Q-diffusion~\cite{li2023qdiffusion} expanded on PTQ4DM and successfully quantized diffusion models trained on high-resolution datasets into INT8 and INT4 data values. PTQD~\cite{he2023ptqd} introduced quantization noise correction methods to further improve the tradeoff between efficiency improvement and output quality degradation. Q-DM~\cite{qdm} quantized diffusion models to lower bitwidth such as INT2 or INT3. However, Q-DM was only evaluated on low-resolution datasets. TDQ~\cite{so2023temporaldynamicquantizationdiffusion} quantized both the weights and activations of diffusion models to INT4 by dynamically adjusting the quantization interval based on time step information. We quantitatively compare our quantization method over Q-diffusion, which is a stronger state-of-the-art baseline in this context, since our goal in this work is to compare the impacts of the same-bitwidth floating-point vs integer quantization on diffusion models. More advanced techniques such as~\cite{qdm,so2023temporaldynamicquantizationdiffusion,he2023ptqd} complement our method to further improve the image generation quality of quantized models.      

\section{Conclusion}
Our work focuses on quantizing diffusion models to achieve high-performance inference. We are the first to quantize the weights of diffusion models to FP8 or FP4 values, and the activations to FP8. We demonstrate that the task performance degradation on unconditional and text-to-image generation caused using our floating-point quantization is significantly better than that of using state-of-the-art integer quantization at the same bitwidth. We also present a better methodology and metric to evaluate the quantization effects on diffusion models. For future work, we aim to explore methods that leverage the sparsity introduced by our quantization method to further optimize diffusion model inference.
\clearpage

\bibliography{reference.bib}
\bibliographystyle{plain}

\end{document}